\documentclass[11pt]{article}

\usepackage[final]{acl}

\usepackage{times}
\usepackage{latexsym}
\usepackage{hyperref}
\usepackage[T1]{fontenc}
\usepackage[utf8]{inputenc}
\usepackage{microtype}
\usepackage{inconsolata}
\usepackage{graphicx}
\usepackage{listings}
\title{
An End-to-End Ukrainian RAG for Local Deployment. \\ Optimized Hybrid Search and Lightweight Generation
}

\author{Mykola Trokhymovych \\
  Pompeu Fabra University\\
  \texttt{mykola.trokhymovych@upf.edu} \\\And
  Yana Oliinyk \\
  Independent Researcher\\
  \texttt{oliinykyana@gmail.com} \\\And
  Nazarii Nyzhnyk \\
  Independent Researcher\\
  \texttt{nazar.nyzhnyk@gmail.com} \\}

\begin{document}
\maketitle
\begin{abstract}
This paper presents a highly efficient Retrieval-Augmented Generation (RAG) system built specifically for Ukrainian document question answering, which achieved 2nd place in the UNLP 2026 Shared Task.\footnote{\url{https://github.com/unlp-workshop/unlp-2026-shared-task}} Our solution features a custom two-stage search pipeline that retrieves relevant document pages, paired with a specialized Ukrainian language model fine-tuned on synthetic data to generate accurate, grounded answers. Finally, we compress the model for lightweight deployment. Evaluated under strict computational limits, our architecture demonstrates that high-quality, verifiable AI question answering can be achieved locally on resource-constrained hardware without sacrificing accuracy. Code is available at: \url{https://github.com/trokhymovych/unlp-2026-shared-task}.
\end{abstract}

\section{Introduction}

Large Language Models (LLMs) have emerged as universal tools, demonstrating remarkable capabilities across a wide range of natural language processing tasks~\cite{minaee2025largelanguagemodelssurvey}. While they encode vast amounts of information within their billions of parameters, this internal knowledge is strictly limited by their training data. LLMs become unreliable when tasks require highly specific knowledge - like texts with information that lie beyond models' training corpus or recent facts~\cite{li-etal-2025-knowledge-boundary}. In these scenarios, they frequently fall back to so-called hallucinations to fill the gaps in their knowledge.

To bridge this gap, Retrieval-Augmented Generation (RAG) has become the definitive framework~\cite{10.5555/3495724.3496517}. By conditioning the generative process on information retrieved from external, domain-specific databases, RAG grounds the LLM in verifiable facts, which aligns with broader efforts in automated fact-checking and knowledge verification~\cite{thorne-vlachos-2018-automated,10.1145/3459637.3481961}. Implementing this, however, adds significant complexity to the data extraction and indexing pipeline. Furthermore, it causes a critical paradigm shift: the precision of the search becomes more important than the raw power of the generative model. Because LLMs rely on the provided context, retrieving incorrect information directly results in an incorrect answer~\cite{niu-etal-2024-ragtruth}.

\begin{figure}
  \centering
  \includegraphics[width=1\linewidth]{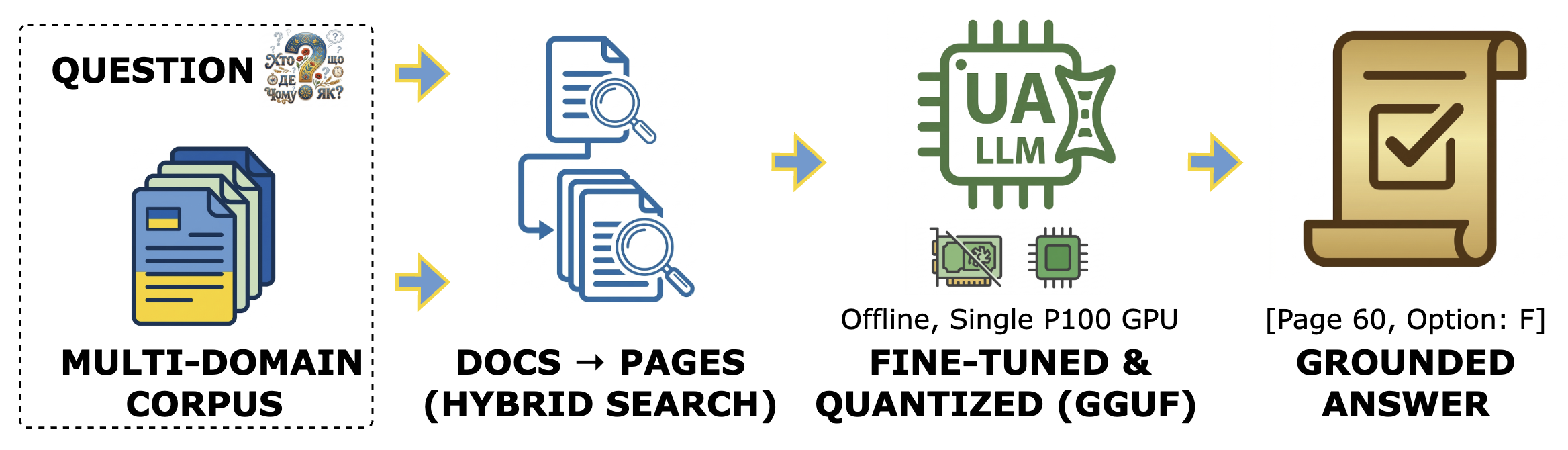}
  \caption{An end-to-end Ukrainian RAG for question answering on local deployment.}
  \label{fig:teaser}
\end{figure}

Moreover, the RAG paradigm inherently requires processing extensive retrieved contexts. At the same time, most language models are heavily optimized for English~\cite{10.1038/d41586-025-03891-y}. Applying these models to low- and medium-resource languages like Ukrainian results in a significant decrease in computational efficiency. Default tokenizers fragment Cyrillic text into significantly more subword pieces than Latin text~\cite{10.3389/frai.2025.1538165}. This exhausts memory limits, significantly slows inference, and degrades accuracy.

A recent UNLP 2026 Shared Task provided a standardized benchmark for these exact challenges, continuing the community's ongoing efforts to advance Ukrainian NLP and information integrity~\cite{kyslyi-etal-2025-unlp,akhynko-etal-2025-hidden}. Participants were required to build a system capable of answering multiple-choice questions in Ukrainian, grounding each answer to a specific document and page within a custom, multi-domain corpus. Additionally, the organizers imposed strict hardware constraints: entirely offline inference on a single P100 GPU within a 9-hour limit. In practice, this meant the solution could not rely on third-party LLM providers and had to be both memory-efficient and fast.

This paper presents our solution to the UNLP 2026 Shared Task (see Figure~\ref{fig:teaser}), which achieved 2nd place on the final leaderboard. Our main contributions are:
\begin{enumerate}
\item A specialized two-stage (document and page-level) hybrid retrieval system.
\item A methodology for generating synthetic Ukrainian QA datasets for model fine-tuning.
\item A customized LLM engineered for high-speed local inference for question answering and grounding.
\end{enumerate}

\section{Related Work}
\subsection{Retrieval-Augmented Generation}

Modern RAG architectures have converged on a multi-stage retrieval approach to balance latency with precision~\cite{gao2024ragsurvey}. These pipelines typically begin with a broad search to identify candidate documents, using fast bi-encoders~\cite{reimers-2019-sentence-bert}, sparse models such as BM25~\cite{Robertson1994OkapiAT}, or a combination of both. Once a candidate set is retrieved, a more computationally intensive cross-encoder reranker is applied to refine the results and ensure high relevance~\cite{DBLP:journals/corr/abs-1901-04085}.

Recent studies suggest that hybrid retrieval, which fuses dense semantic embeddings with lexical search, consistently outperforms single methods by capturing both conceptual meaning and exact keyword matches~\cite{akarsu2026bm25correctiveragbenchmarking}. A common technique for this fusion is Reciprocal Rank Fusion (RRF), an unsupervised method that merges disparate ranked lists by summing the reciprocal of document ranks~\cite{10.1145/1571941.1572114}.

Prior to retrieval, processing large documents necessitates chunking~\cite{gao2024ragsurvey}, a step that often causes individual segments to lose their thematic connection to the original source. To address this, contextual embeddings can be utilized to preserve document-wide intent within each fragment~\cite{eslami2026diffusionpretraineddensecontextualembeddings}.

\subsection{Language Models for Ukrainian}

Modern Large Language Models (LLMs) can effectively handle many languages~\cite{gemmateam2025gemma3technicalreport,ustun-etal-2024-aya}, but they still favor high-resource languages, simply because they dominate the text used to train them. Similar challenges have been previously observed across various NLP tasks, from evaluating text readability to maintaining knowledge integrity and question answering~\cite{10.1145/3580305.3599823,trokhymovych-etal-2024-open,zhang-etal-2023-dont}. As a result, models struggle with medium-resource languages like Ukrainian. Standard tokenizers break Ukrainian words into too many tokens~\cite{10.3389/frai.2025.1538165}. This wastes space, makes processing slower, and hurts the model's overall performance.

To address these gaps, recent work has introduced models specifically optimized for Ukrainian, such as MamayLM~\cite{MamayLM}. Built on the Gemma 3 12B architecture, it underwent continual pre-training on a large, pre-filtered dataset using a combination of data mixing and model merging to gain exceptional Ukrainian cultural and linguistic proficiency. Despite its 12B-parameter size, MamayLM matches or exceeds the performance of significantly larger models, including Llama 3.1 70B, on Ukrainian-specific tasks~\cite{MamayLM}.

\subsection{LLM Adaptation and Deployment}

LLMs are typically developed as general-purpose models capable of performing a wide range of tasks. However, they can benefit from fine-tuning for specialized applications. Given their massive size, traditional fine-tuning is often computationally impossible under hardware constraints. To address this, previous work introduced LoRA (Low-Rank Adaptation), which only updates small, low-rank matrices that serve as adapters to original model weights modifying key layers such as the attention layers~\cite{DBLP:journals/corr/abs-2106-09685}. This approach enables the efficient adaptation of LLMs for specific tasks such as grounded question answering.

There are also several approaches that enable efficient inference under hardware constraints. In particular, we employ quantization via the GGUF format and the llama.cpp library~\cite{llama_cpp}. Quantization reduces the numerical precision of model weights, which significantly decreases memory usage and increases throughput with only a moderate impact on accuracy~\cite{10628367}. The GGUF format facilitates this process by storing quantized weights alongside essential metadata, enabling efficient loading and execution across diverse hardware configurations.

\section{System Architecture}
This section describes our final solution for the UNLP 2026 Shared Task. Built on a RAG framework, our system employs a modular pipeline designed to preserve document structure during processing, perform relevant context retrieval across multi-domain corpora, and generate grounded answers to the questions.

\subsection{Data Preparation}
To convert raw PDF documents into a format suitable for Large Language Models (LLMs), we utilize \texttt{pymupdf4llm}\footnote{\url{https://pymupdf.readthedocs.io/en/latest/pymupdf4llm/}} to perform layout-aware extraction. 
This method converts document pages into Markdown, which preserves structural elements such as tables and headers. 

Only for lexical search, we implement a custom pre-processing function using \texttt{pymorphy3}\footnote{\url{https://pypi.org/project/pymorphy3/}} to perform lemmatization and tokenization. Additionally, we filter out a comprehensive list of Ukrainian stop words\footnote{\url{https://github.com/skupriienko/Ukrainian-Stopwords}} and generic artifacts to improve the signal-to-noise ratio during the retrieval. This function is applied to both the query and the corpus for lexical search. 

\subsection{Hybrid Retrieval Pipeline}
Our retrieval strategy is a two-stage process that first identifies the correct document and then finds the most relevant pages within that document for each question.

\subsubsection{Document-Level Retrieval.}
The initial search narrows the candidate documents to the single most relevant file. We employ a hybrid scoring mechanism that combines:
\begin{itemize}
\item Dense Retrieval: We use Perplexity embeddings (\texttt{pplx-embed-context-v1-0.6b}\footnote{\url{https://huggingface.co/perplexity-ai/pplx-embed-context-v1-0.6b}}), to capture the semantic similarity between the query and the document content. Model choice is motivated by strong performance on benchmarks and compact 0.6B size
\item Sparse Retrieval: We utilize the \texttt{BM25Okapi} algorithm.
\end{itemize}

To perform the document search, we represent a query as the question concatenated with its corresponding answer options. To build the vector representation of each document, we embed its first 300 characters and use cosine similarity for ranking. For lexical search, we index each document using its full preprocessed text and use the BM25 score for ranking.

We found that for the majority of questions, the top-ranked item matches for both sparse and dense retrieval and is the correct document. For others, the correct document appears in the top two retrieved documents for at least one approach. Therefore, if the top results match, we return that document. Otherwise, we select the top two documents from each approach and use \texttt{jina-reranker-v3}\footnote{\url{https://huggingface.co/jinaai/jina-reranker-v3}} to find the best match. For the reranker, each document is represented by concatenating its first 300 characters with its best BM25 snippet. This algorithm allowed us to achieve near-perfect performance for document retrieval.

\subsubsection{Page-Level Retrieval}
Once a document is selected, we split it into small parts using syntactic chunking based on Markdown structure. Specifically, we ensure that each chunk is no longer than 500 characters, has a 10\% overlap with adjacent chunks, and corresponds to only one page.

Embeddings for each chunk are calculated using the same contextual model as for document-level retrieval. Due to limited resources, we encode chunks in batches of 5 with an overlap of 2 to preserve context. To obtain the final embedding for items in the overlap, we average the results from both batches.

Embeddings are used to identify the chunks ranking via cosine similarity with the question vector. Additionally, we generate a separate ranking based on the BM25 score. We then apply Reciprocal Rank Fusion (RRF) to merge the vector and BM25 scores ranking. Finally, we use a custom cross-encoder based on \texttt{BAAI/bge-reranker-v2-m3} to rerank the top-8 candidates, providing a final list of the most relevant chunks for generation.
The model was chosen for its strong performance on benchmarks and simple architecture that allows for efficient fine-tuning and inference on a P100 GPU. 

\subsection{Answer Generation and Grounding}
Based on a ranked list of chunks, we build the context out of the full text of the top-3 relevant pages. 
We pass the context, question, and options to the fine-tuned MamayLM-12B generative model (see the prompt in Appendix~\ref{sec:app_prompt}). The model is specifically fine-tuned via Low-Rank Adaptation (LoRA) to perform the dual task of generating an answer with the page number (e.g., ``A 2'') to ground the response. To optimize the model for deployment in resource-constrained environments (such as Kaggle's P100 GPU), we utilize GGUF 4-bit quantization via the \texttt{llama-cpp-python} library.

\section{Experimental Setup}
\subsection{Data}
The competition provides a development dataset consisting of 461 questions and 41 documents from two distinct domains. We use this dataset as the primary data for local evaluation of our solution. It should be noted that documents in the corpus vary significantly in length, with some reaching more than 100 pages.

The Shared Task is structured as a code competition, where solution inference is performed on a hidden test set. The leaderboard is split into public (27\% of the test data) and private (73\% of the test data) sections, with the latter defining the final team ranking. The only information available regarding the test data is that the corpus consists of more than 240 files from a new secret domain and contains a significantly larger number of questions. 

To expand the training dataset for custom reranker and foundational LLM fine-tuning, we developed an automated pipeline that synthesizes multiple-choice questions (MCQs) directly from source PDF documents. The pipeline processes documents page by page, first evaluating whether a page contains factual content suitable for question generation. It is explicitly instructed to skip non-factual sections, such as title pages, tables of contents, abbreviation lists, and general introductions.

For each page, the model generates up to ten MCQs. Each question includes exactly six answer options (A through F) with a single correct answer and reference to the document corpus reproducing the structure of the original development data. To ensure each generated question is self-contained and answerable without referencing the source document, the prompt enforces a critical constraint: every question must explicitly state the relevant entity name (e.g., the specific sport or drug).

Generation was performed using OpenAI's \texttt{gpt-4o-mini} model, configured with a temperature of 0.7. The complete prompt used for this generation is provided in Appendix~\ref{sec:app_prompt}. As a result, this process yielded an additional synthetic dataset of about 7,000 questions.

\subsection{Reranker Model Fine-Tuning}
The custom reranker is trained using the \texttt{BAAI/bge-reranker-v2-m3} model with a hybrid loss objective that combines Listwise Cross-Entropy (weight 1.0) and Pointwise Binary Cross-Entropy (BCE) (weight 0.4) to optimize document chunk ranking. For each query, the model processes a group of candidate chunks and computes a relevance score for each. 
The model is trained for 5 epochs with a batch size of 2 and gradient accumulation over 4 steps. Final model selection is based on Top-1 Accuracy performance on a hold-out validation subset.

\subsection{Generative Model Fine-Tuning}
For the generation stage, we fine-tune the \texttt{MamayLM-Gemma-3-12B-IT-v1.0} model using Low-Rank Adaptation (LoRA) with a fixed prompt structure (see Appendix~\ref{sec:app_prompt}). The training data is processed using a syntactic masking strategy, where the loss is calculated exclusively on the model's generated answer (letter and page number) by masking the prompt tokens. To accommodate long-context documents, the implementation includes a dynamic truncation mechanism with a maximum sequence length of 4000 tokens and utilizes Flash Attention 2 and gradient checkpointing for efficiency. The model is trained using the AdamW 8-bit optimizer and a cosine learning rate scheduler, with a custom evaluation suite that independently tracks accuracy for both the multiple-choice answer and the cited page number. We fine-tune the model for two epochs, first using synthetic data and subsequently on the competition development data, to ensure maximal adaptation of the model to the testing set. 

\section{Results}

\subsection{Evaluation Metrics}
The evaluation metric is a weighted average that assesses three specific components of the model's output across $N$ questions. Half of the total score ($0.5$) is determined by the accuracy of the multiple-choice answers, $a_i$, where the model receives a point only for an exact match with the ground truth. 

The remaining half of the score focuses on the retrieval reference and is split equally ($0.25$ each) between the document ID, $d_i$, and the page proximity, $p_i$. The document score is binary, rewarding the model for identifying the correct file. The page proximity score is more granular; it calculates the distance between the predicted and true page numbers relative to the total number of pages in the document. Crucially, this proximity credit is only awarded if the correct document was identified first ($d_i = 1$). This structure penalizes complete retrieval failures while providing partial credit for "near misses" on specific page locations within the correct document.

\subsection{Competition Results}
In the final ranking, we achieved second place with a score of 0.942 on the private test set and 0.920 on the public set. Our pipeline successfully processed the hidden test corpus, exactly meeting the strict 9-hour compute limit on a single P100 GPU. Local validation scores on the development set closely matched these final results, demonstrating strong generalization to the unseen domain. Finally, our two-stage hybrid retrieval achieved near-perfect accuracy for document identification and a page-level recall@3 of 0.92 on the development dataset.

\section{Conclusion}
We presented an efficient, end-to-end RAG system for Ukrainian QA, achieving 2nd place in the UNLP 2026 Shared Task. By combining a two-stage hybrid retrieval pipeline, fine-tuning based on synthetic data, and a quantized \texttt{MamayLM-12B model}, we demonstrated that accurate, grounded question answering is viable on local, resource-constrained hardware.

\section*{Limitations}
Our approach presents several limitations, primarily related to the strict hardware constraints of the shared task. First, to minimize computational overhead, we bypass the processing of images and charts, thereby omitting visual context that could otherwise enhance retrieval and generation accuracy. Second, a marginal fraction of complex PDF pages failed during the layout extraction phase. Although Optical Character Recognition (OCR) could recover this text, the additional processing time outweighed the benefits, given the low incidence rate. Finally, we relied on standard 4-bit GGUF quantization to ensure reliable execution on the provided legacy GPU architecture (P100). Exploring alternative, state-of-the-art quantization techniques could potentially yield further performance improvements if deployed on more modern hardware.

\section*{Ethical Considerations}
While our system advances Ukrainian NLP, the foundational models and synthetic data pipeline may inherit biases from their pre-training corpora. Furthermore, despite RAG's grounding mechanisms, the risk of hallucination persists. Given the inclusion of sensitive domains like pharmaceutical data, this system requires human oversight for real-world applications. On the other hand, our focus on fully offline, local inference ensures strict data privacy and reduces the environmental footprint of deployment.

We acknowledge the use of AI tools in the preparation of this manuscript. As the authors are non-native English speakers, Google Gemini and Grammarly were used to correct grammar and refine language, improving readability. Additionally, a generative AI model was utilized to create visual elements for the paper's teaser image. Ethically, these tools were not used to generate scientific claims, experimental data, or core ideas, ensuring all intellectual contributions remain solely those of the authors.

\section*{Acknowledgments}
The work of Mykola Trokhymovych is supported by the Google PhD Fellowship and 
MCIN/AEI /10.13039/501100011033 under the
Maria de Maeztu Units of Excellence Programme
(CEX2021-001195-M).

\bibliography{custom}

\appendix
\section{Prompts}
\label{sec:app_prompt}

\subsection{Answer Generation Prompt}
\label{sec:app_gen_prompt}
The following prompt is used for final answer generation with page grounding (Figure~\ref{prm:generation}).

\begin{figure}[ht]
\begin{lstlisting}
Context (excerpts from PDF files - each excerpt is
separated by ``` characters and contains a page
number enclosed in []):
```
Page: [<page_number_1>]
<page_text_1>
```
```
Page: [<page_number_2>]
<page_text_2>
```
```
Page: [<page_number_3>]
<page_text_3>
```

Question: <question>
Options:
A: <option_A>
B: <option_B>
C: <option_C>
D: <option_D>
E: <option_E>
F: <option_F>
Instructions:
- Answer the Question using the Context.
- Return the letter of the correct answer (A B C D E F)
  and the page number where the information was found,
  separated by a space (e.g., A 1).
- Think carefully; first eliminate the obviously
  irrelevant options.
\end{lstlisting}
\caption{Prompt template for answer generation with grounding (translated from Ukrainian).}
\label{prm:generation}
\end{figure}

\subsection{Synthetic Question Generation Prompt}
\label{sec:app_synth_prompt}
The following prompt is used for synthetic MCQ generation (Figure~\ref{prm:generation_2}).

\begin{figure}[ht]
\begin{lstlisting}
SYSTEM:
You are an expert Ukrainian-language exam question writer.

DOMAIN CONTEXT:
{domain_description}

YOUR TASK:
You will receive the text of a single page from a Ukrainian PDF document. You must:

1. IDENTIFY the specific subject: the exact sport name
   (e.g. "strongman", "sambo") or drug name
   (e.g. "retabolil", "fervex"). This is the ENTITY NAME.
2. DECIDE whether this page contains specific factual
   content suitable for question generation (specific
   rules, dosages, penalties, contraindications, etc.).
   SKIP pages that are tables of contents, title pages,
   abbreviation lists, or general introductions.   
3. If suitable, generate up to 10 MCQs that:
   - Are written entirely in Ukrainian.
   - CRITICAL: Every question MUST explicitly name the
     ENTITY NAME. Generic questions are FORBIDDEN.
   - Are answerable ONLY from the provided page text.
   - Have exactly 6 options (A-F), one correct answer,
     and 5 plausible distractors.
   - NEVER use quotation marks or braces.
   - Match the style of: {few_shot_examples}
4. If NOT suitable, return an empty questions list.

RESPONSE FORMAT (strict JSON):
{
  "entity_name": "<sport or drug name>",
  "questions": [
    {
      "question": "...",
      "A": "...", "B": "...", "C": "...",
      "D": "...", "E": "...", "F": "...",
      "correct_answer": "A"
    }
  ]
}

Return ONLY valid JSON. "correct_answer" must be one of: A, B, C, D, E, F.
If not suitable: {"entity_name": "", "questions": []}

USER:
Domain: {domain}
Document page text: {page_text}
\end{lstlisting}
\caption{Prompt template used for synthetic question generation. Note: examples appear in Ukrainian in the actual prompt.}
\label{prm:generation_2}
\end{figure}

\end{document}